\documentclass[preprint,12pt]{elsarticle}

\usepackage{amssymb}

\usepackage{lineno}

\usepackage{amsmath}
\usepackage{graphicx}
\usepackage{multirow}
\usepackage{array}
\usepackage{hyperref}
\hypersetup{
	colorlinks=true,
	linkcolor=black,
	filecolor=blue,      
	urlcolor=blue,
	citecolor=black,
}

\usepackage{booktabs}
\usepackage{setspace}
\usepackage{cases}
\usepackage{pifont}
\usepackage{array}
\usepackage{subfigure}
\usepackage{bm}

\usepackage[utf8]{inputenc}

\journal{Name of journal}

\begin{document}

\begin{frontmatter}



\title{Focus-Enhanced Scene Text Recognition with Deformable Convolutions}


\author{Linjie Deng,
	Yanxiang Gong,
	Xinchen Lu,
	Xin Yi,
	Zheng Ma,
	Mei Xie\textsuperscript{*}}

\address{School of Information and Communication
	Engineering, University of Electronic Science and Technology of China, No.2006, Xiyuan Ave,
	Chengdu, China.
	
	*Mei Xie is the corresponding author.}

\begin{abstract}
Recently, scene text recognition methods based on deep learning have sprung up in computer vision area. The existing methods achieved great performances, but the recognition of irregular text is still challenging due to the various shapes and distorted patterns. Consider that at the time of reading words in the real world, normally we will not rectify it in our mind but adjust our focus and visual fields. Similarly, through utilizing deformable convolutional layers whose geometric structures are adjustable, we present an enhanced recognition network without the steps of rectification to deal with irregular text in this work. A number of experiments have been applied, where the results on public benchmarks demonstrate the effectiveness of our proposed components and shows that our method has reached satisfactory performances. The code will be publicly available at \url{https://github.com/Alpaca07/dtr} soon.

\end{abstract}


\end{frontmatter}



\section{Introduction}
\label{intro}
\begin{figure}
	\centering
	\begin{center}
		\subfigure[]{\includegraphics[width=0.7in]{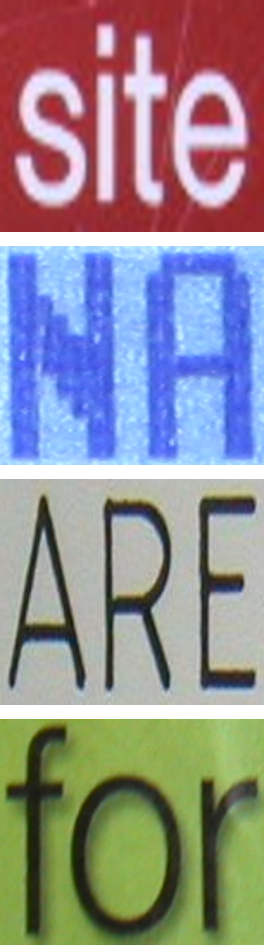}}
		\subfigure[]{\includegraphics[width=0.7in]{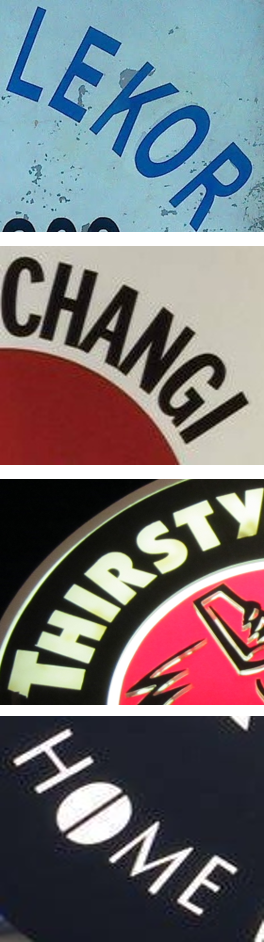}}
		\subfigure[]{\includegraphics[width=0.7in]{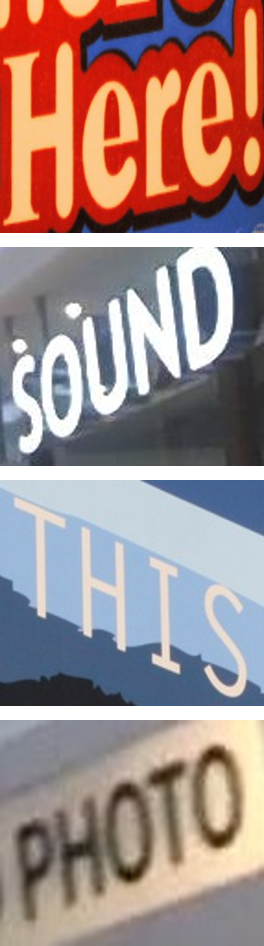}}
		\subfigure[]{\includegraphics[width=0.7in]{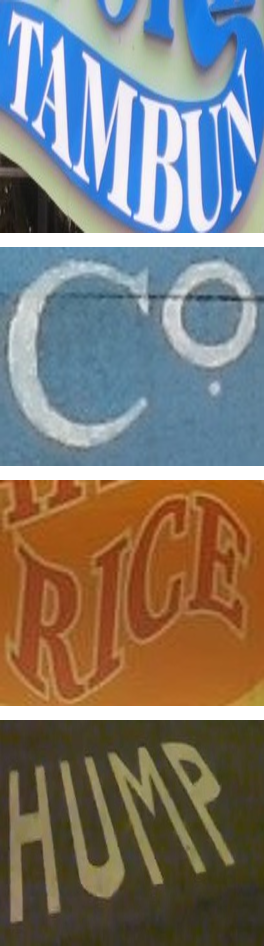}}
	\end{center}
	\caption{The examples of images with regular and irregular text from public benchmarks. (a) Regular text. (b) Curved Text. (c) Tilted text. (d) Other kinds of irregular text.}
	\label{examples}
\end{figure}

Text in scene images usually contains a large amount of semantic information, thus text recognition plays an important role in the field of computer vision. With the advent of deep learning, methods for recognition have made great progresses\cite{crnn,mjsynth1,mjsynth2,synth}. These techniques achieve great performances on regular text images, but they are not expert in treating irregular text images owing to the fixed geometric structures of the layers in the modules. Unfortunately, irregular text is also very common in the wild, as illustrated in Figure \ref{examples}. Therefore, some predecessors\cite{liu2016star,cheng2017focusing,sun2018textnet} utilized rectification networks or attention mechanism to mitigate this issue. Yao \textsl{et al.}\cite{yao2014unified} presented to use a dictionary to do error correction on the recognition results to handle multi-oriented text images. Luo \textsl{et al.}\cite{moran} added a multi-object rectification network before the recognition network. Shi \textsl{et al.}\cite{shi2016robust} put forward a spatial transformer network\cite{stn} to automatically rectify the word images. Gupta \textsl{et al.}\cite{synth} proposed a method to regard the whole word as a class that will ignore the arrangement of the characters. Lyu \textsl{et al.}\cite{lyu2018mask} propose an end-to-end learning procedure to handle text instances of irregular shapes. These methods are highly effective and greatly alleviated the problem of irregular text recognition, yet most of them tend to rectify the images from different perspectives. That may lead to more manual designs like requirements of preprocesses and increment of network complexity.

In most recognition networks, standard convolutional layers that possess receptive fields with a fixed rectangular shape are utilized. These layers have a certain effect, but they may not best suit the text area as there are redundant background noises, as shown in Figure \ref{examples}. For a better performance, especially on irregular text images, we propose a focus-enhanced text recognition method in which the operation of rectifying images is needless in this paper. Imagine that we are reading text in the real world, if the words are arranged irregularly, usually we will not turn our head to fit the arrangements of them or try to rectify their shapes in our minds. Generally we simply make our focus move along with the text line and change our visual field in the minds. Hence we intend to give the network a capability to focus on the text area and extract the feature precisely. Recently, deformable convolutional layers which are able to learn additional offsets on each sample location of the convolutional kernels from the images are proposed by Dai \textsl{et al.}\cite{deconv}, which inspires us to make the layers change their structures. In our work, deformable layers that are able to adjust the receptive field to cover the region of interest better have been applied in order to enhance the focus of text recognition networks. We integrate our components with CRNN\cite{crnn}, which is widely utilized as a baseline model. To assess our method and confirm the effectiveness, we carry out some ablation experiments and comparisons with other methods on several public benchmarks. The results demonstrate that our proposed method can achieve a competitive performance both on regular and irregular text benchmarks. All the training and testing codes will be open source soon.

\section{Methodology}
\label{method}

\subsection{Baseline}
In this section, first we review the architecture of CRNN(Convolution Recurrent Neural Networks) \cite{crnn}. To the best of our knowledge, it is the first attempt to integrate CNN(Convolution Neural Networks) and RNN(Recurrent Neural Networks) for scene text recognition. Benefited from RNN who is able to achieve sequence labeling without segmentation, the model realizes end-to-end recognition. The network mainly composes of three modules: convolution, recurrent and transcription. Based on the VGG architectures\cite{vgg}, the convolutional layers consist of convolution and max pooling, and it is responsible for extracting the features of the text into frames. In the recurrent layers, BiLSTM(Bidirectional Long Short-Term Memory) networks\cite{lstm} which utilize a forget gate to avoid the vanishing gradient problem are utilized, and the layers predict the content of each frame. The transcription layer decodes the per-frame predictions into a label sequence.

\subsection{Deformable Modules}

\begin{figure}[t]
	\centering
	\begin{center}
		\includegraphics[width=4in]{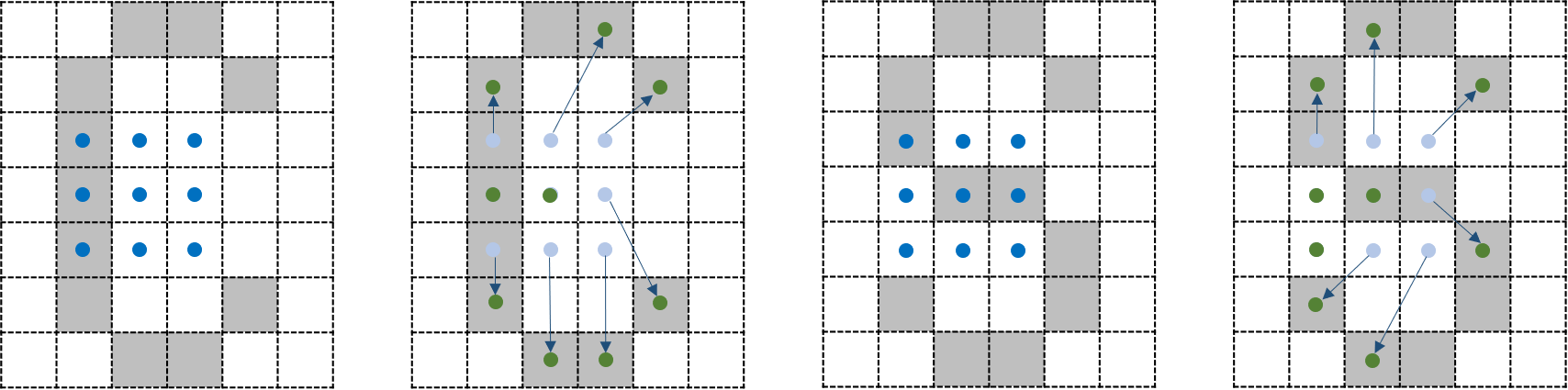}
	\end{center}
	\caption{Indication of $3\times3$ deformable convolution kernel. The blue points indicate the sample locations of a standard convolution kernel. The arrows shows the offsets, and the green points indicate the sampling locations of deformable convolution. }
	\label{deformable_illustrate}
\end{figure}

Although CRNN\cite{crnn} has reached great performances on regular text images, the models still cannot achieve a satisfactory performance when facing irregular text images on account of the background noises and the training data most of which are regular. Enlightened by Dai \textsl{et al.}\cite{deconv}, who present a network in which the convolutional layers can learn to add offsets on each sample location, we consider that it is necessary for the network to change the focus to treat irregular text. Hence we intend to integrate these deformable layers into our baseline model to enhance its focus. The 2D convolution of each location $p_0$ in the image can be expressed as 

\begin{equation}
\bm{y}(\bm{p}_0)=\sum_{\bm{p}_n\in R}^{}\bm{w}(\bm{p}_n)\bm{x}(\bm{p}_0+\bm{p}_n)
\end{equation}
where $x$ represents the input feature map, $w$ represents the weight of sampled values and $R$ defines the receptive field size and dilation. In the defromable convolution, $R$ is augmented with offsets$\{\Delta \bm{p}_n|n=1,...,N\}$

\begin{equation}
\bm{y}(\bm{p}_0)=\sum_{\bm{p}_n\in R}^{}\bm{w}(\bm{p}_n)\bm{x}(\bm{p}_0+\bm{p}_n+\Delta \bm{p}_n)
\end{equation}

\begin{figure}[t]
	\centering
	\begin{center}
		\subfigure[]{
			\begin{minipage}[t]{0.45\linewidth}
				\centering
				\includegraphics[width=2.2in]{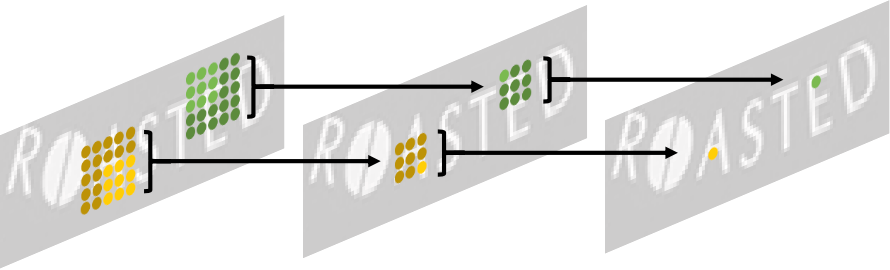}
				\label{standard_visualize1}
			\end{minipage}
		}
		\subfigure[]{
			\begin{minipage}[t]{0.45\linewidth}
				\centering
				\includegraphics[width=2.2in]{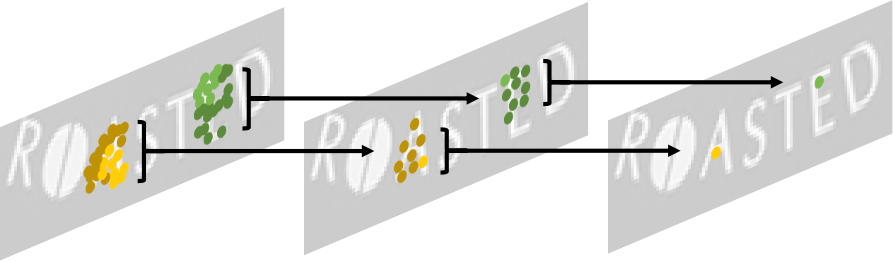}
				\label{deformable_visualize1}
			\end{minipage}
		}
		
		\subfigure[]{
			\begin{minipage}[t]{0.45\linewidth}
				\centering
				\includegraphics[width=2.2in]{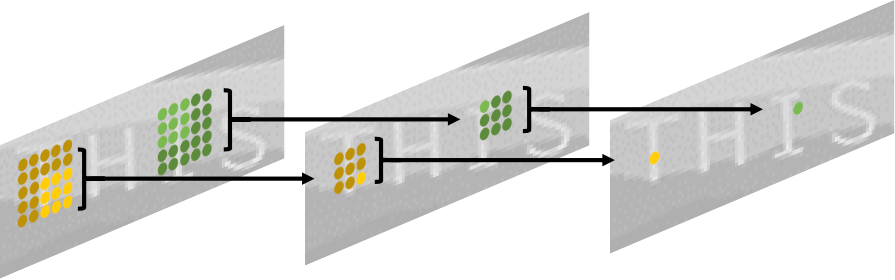}
				\label{standard_visualize2}
			\end{minipage}
		}
		\subfigure[]{
			\begin{minipage}[t]{0.45\linewidth}
				\centering
				\includegraphics[width=2.2in]{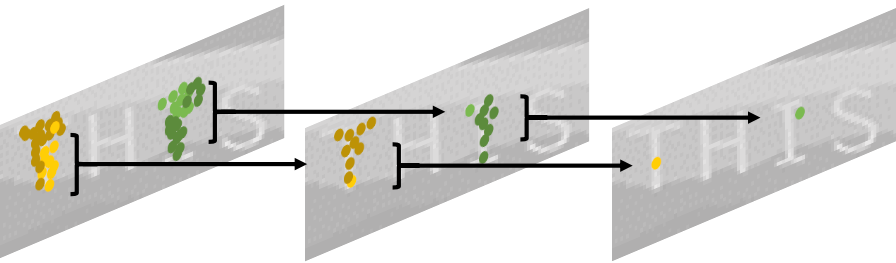}
				\label{deformable_visualize2}
			\end{minipage}
		}
	\end{center}
	\caption{Indication of fixed receptive fields in standard convolution(a)(c) and adaptive receptive fields in deformable convolution(b)(d). In each image triplet, the left shows the sampling locations of two levels of $3\times3$ filters on the preceding feature map, the middle shows the sampling locations of a $3\times3$ filter and the right shows two activation units. Two sets of locations are highlighted according to the activation units.}
	\label{deformable_visualize}
\end{figure}

The offsets on sampling locations can be learned automatically during the training stage. The process of the deformable convolution is indicated in Figure \ref{deformable_illustrate}. It is obvious that the shapes of the receptive fields of the deformable convolution can better focus on the text area, which enables the ability to treat irregular text of our network. However, the replacement is not arbitrary. The shallow layers usually extract some basic information such as the edges, shapes and textures, thus the deformable convolution may be ineffective. And because there should be some space in the image for the receptive fields to drift, the effectiveness could be not obvious as deep layers have a too small size. In summary, we finally replace the convolutional layers in the middle of the network with deformable ones, which will be introduced in details in section \ref{location_impact}. The visualization of the layers in our network is shown in Figure \ref{deformable_visualize}.

\subsection{Network Architecture}
\begin{figure}[h]
	\centering 
	\begin{flushleft}
		\subfigure[]{
			\begin{minipage}[t]{0.58\linewidth}
				\centering
				\includegraphics[height=2.9in]{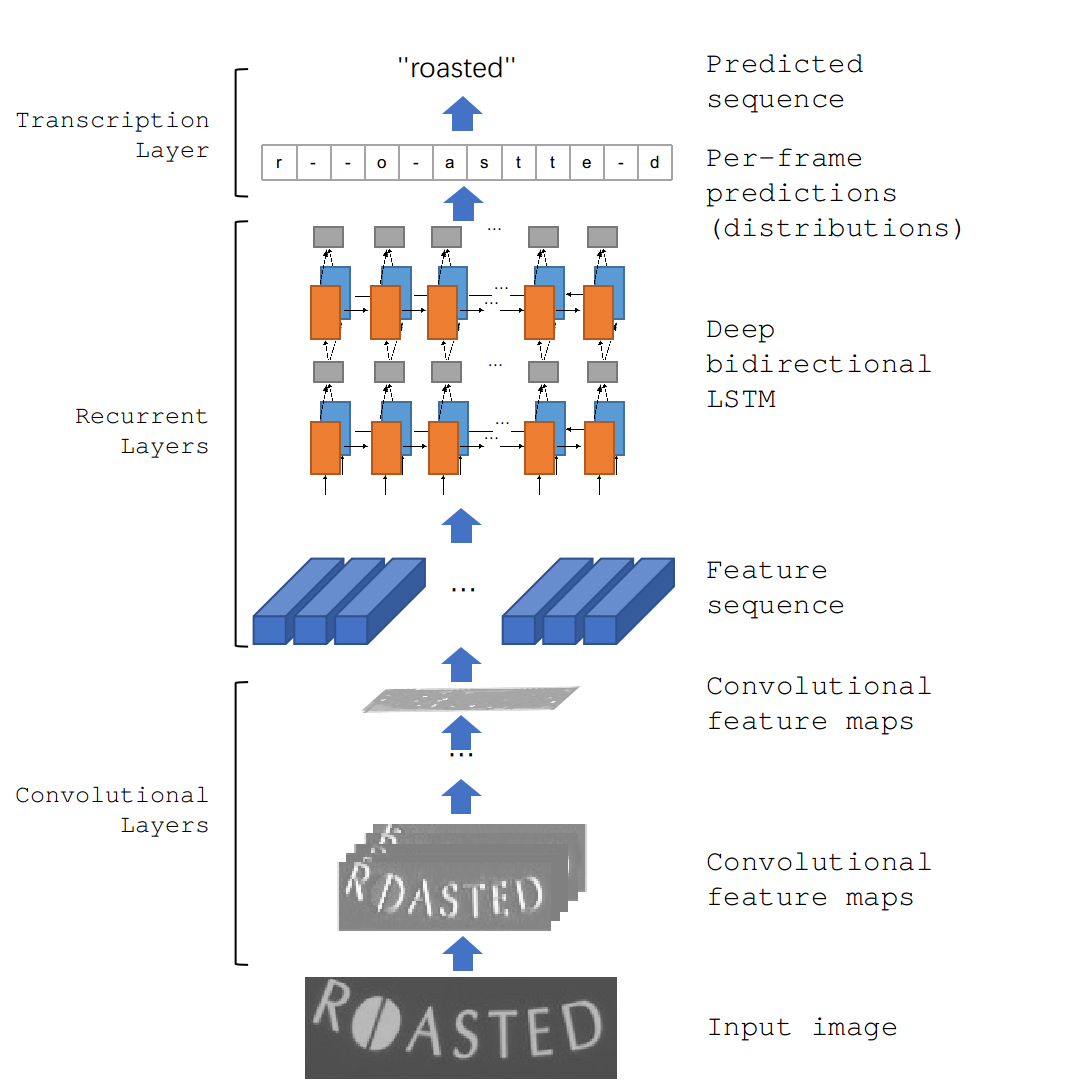}
				\label{pipline_of_crnn}
			\end{minipage}%
		}
		\subfigure[]{
			\begin{minipage}[t]{0.32\linewidth}
				\centering
				\includegraphics[height=2.9in]{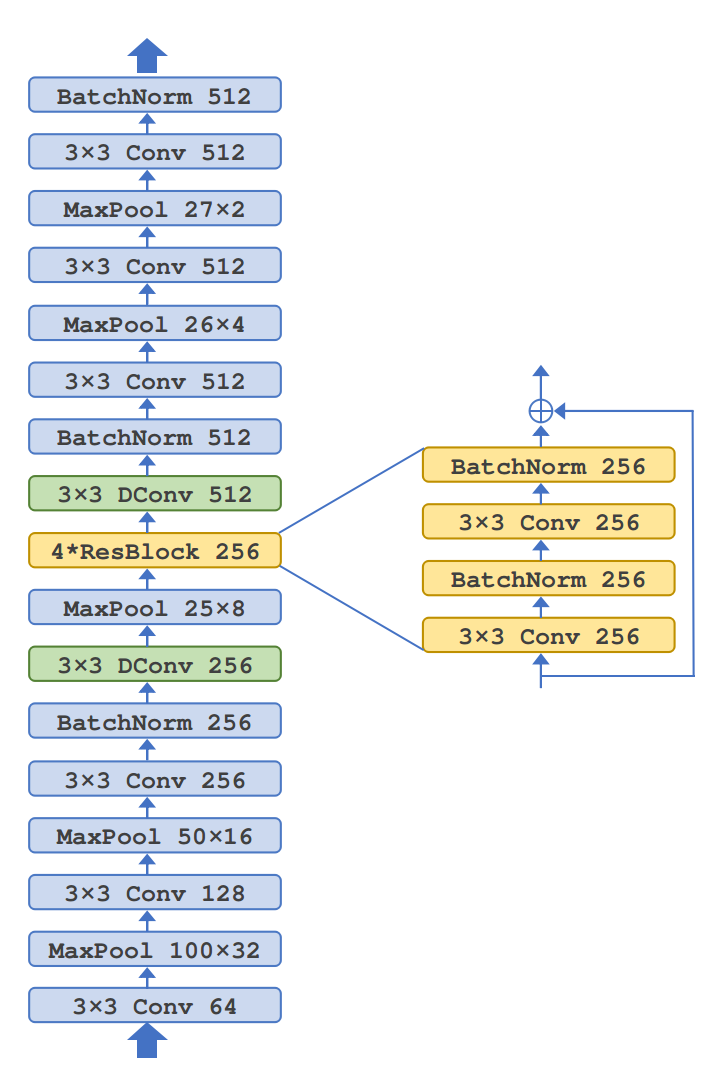}
				\label{convolutional_layers_archeticture}
			\end{minipage}
		}
	\end{flushleft}
	\centering
	\caption{The pipeline of our method(a), the architecture of the convolutional layers in the network and the components of the residual blocks(b). In the figure, "Conv" means a standard convolutional layer with the kernel size and output channel number, "MaxPool" means an adaptive max pooling layer with the output size, "DConv" means a deformable convolutional layer with the kernel size and output channel number and "BatchNorm" means batch normalization operation with the input channel number.}
	\label{architectures}
\end{figure}

The architecture of oue modified network is depicted in Figure \ref{architectures}. Owing that the network becomes more complicated, in order to avoid the problem of vanishing gradient, some residual blocks have been utilized. Each of the residual blocks consists of two $3\times3$ convolutions with a \textsl{skip connection}. And for better engineer implementation, the adaptive max pooling layers are applied in our network. These layers are able to get the kernel size automatically according to the required size of the output feature map with

\begin{equation}
k=in-(out-1)\times floor(\frac{in}{out})
\end{equation}
where $in$ means the input kernel size and $out$ means the output kernel size.
The pipeline of our network is shown in Figure \ref{pipline_of_crnn} and the architecture of the convolutional layers is depicted in Figure \ref{convolutional_layers_archeticture}. All the activation functions are ReLU and we do not modify the recurrent layers and the transcription layer of the baseline model.

\section{Experiments}
\label{experiment}

In the following part, we will describe the implementation details of our method. We use the standard evaluation protocols and run a number of ablations to analyze the effectiveness of the proposed components on public benchmarks.

\subsection{Datasets}
\subsubsection{Training Data}
\label{training_data}	
\noindent\textbf{MJSynth Dataset}\cite{mjsynth1, mjsynth2} includes about 9-million gray synthesized images. The fonts of the text are collected from Google Fonts and the backgrounds are collected from ICDAR 2003\cite{icdar03} and SVT\cite{SVT} datasets. Text samples are obtained through font, border and shadow rendering. Then after coloring and distortion, the samples will be blended into natural images with some noises.

\noindent\textbf{SynthText in the Wild Dataset}\cite{synth} includes about 7-million colored synthesized images. The dictionary is collected from SVT dataset\cite{SVT} and the background images are from Google Image Search. To synthesize text images, a text example will be rendered and then blended into a contiguous region of a scene image according to the semantic information. 

\subsubsection{Testing Data}
To confirm the effectiveness of our method, five benchmarks have been utilized in the experiments. Each of them will be introduced below.

\noindent\textbf{TotalText(Total)}\cite{totaltext} dataset consists of 1255 training images and 300 testing images with more than 3 different text orientations: horizontal, multi-oriented, and curved. There are totally about 2,000 word images.

\noindent\textbf{ICDAR 2013(IC13)}\cite{icdar13} dataset consists of 229 training and 233 testing images which contains about 1,000 words. The images were captured by user explicitly detecting the focus of the camera on the text content of interest, so most of the text is regular arranged horizontally. 

\noindent\textbf{ICDAR 2015(IC15)}\cite{icdar15} dataset consists of 1000 training and 500 testing images with about 4,000 words. The images were collected without taking any specific prior attention, thus the word images may be blurred and not arranged horizontally.

\noindent\textbf{Street View Text(SVT)}\cite{SVT} dataset consists of 353 images in which about 600 words were labeled as testing data. The images were collected from Google Street View and many of them are severely corrupted by noise and blur, or have very low resolutions.

\noindent\textbf{IIIT 5K-Words(IIIT5K)}\cite{IIIT5K} dataset contains 3,000 cropped word test images. The dataset is harvested from Google image search.

In these datasets, Total\cite{totaltext} and IC15\cite{icdar15} are commonly used to test the ability to recognize irregular text images, and the others are mainly to test recognition of regular ones. 

\subsection{Implementation Details}
The network is trained only with the synthetic text images mentioned in section \ref{training_data}, and no real data is involved. All of the input images are resized to $200\times64$. The loss function is Connectionist Temporal Classification (CTC) loss proposed by Graves \textsl{et al.}\cite{ctc}. The optimizer of the network is SGD, the batch size is set to 64 and the learning rate is 0.00005. The network is trained for 8 epochs which costs 3 days. The proposed method is implemented by PyTorch\cite{pytorch}. All experiments are carried out on a standard PC with Intel i7-8700 CPU and a single Nvidia TITAN Xp GPU.

\subsection{Ablation Experiments}

\subsubsection{The Impacts of Proposed Components}
\begin{figure}[htbp]
	\centering
	\begin{center}
		\begin{minipage}[t]{1\linewidth}
			\centering
			\includegraphics[width=4.5in]{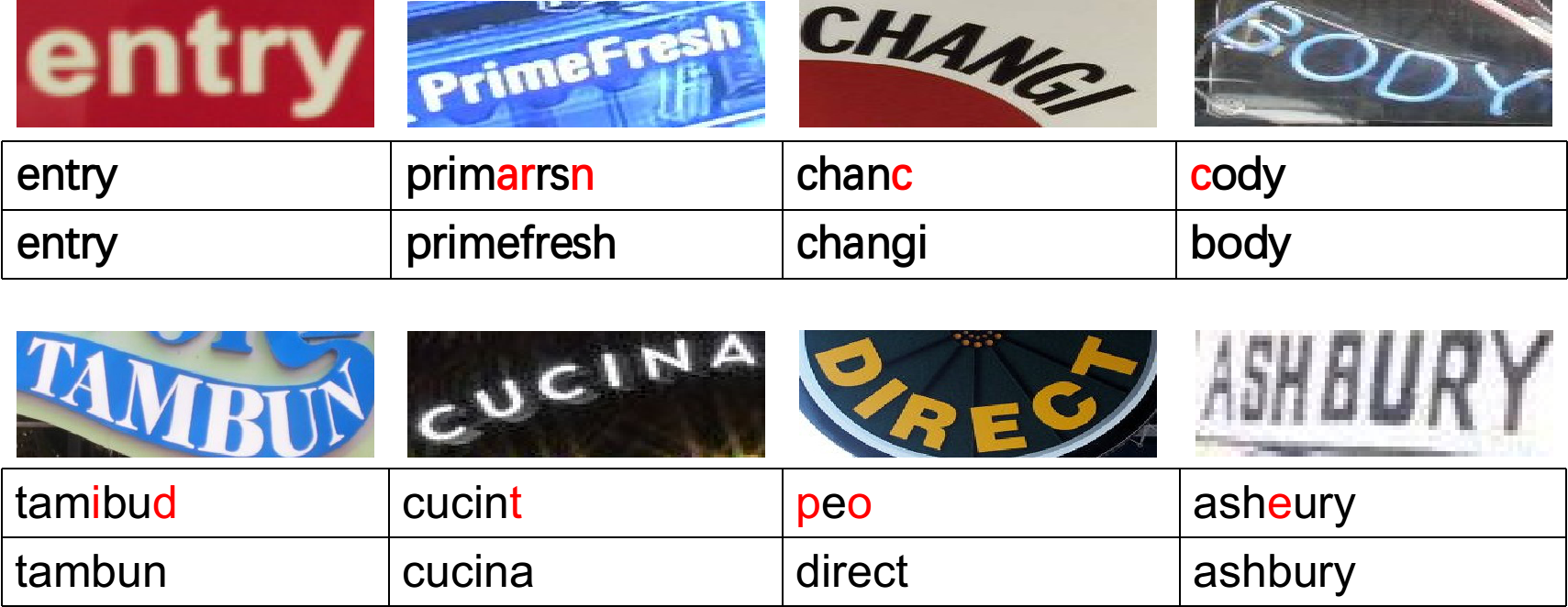}
		\end{minipage}
	\end{center}
	\centering
	\caption{The recognition results of our baseline model(up) and our method(down). The red characters are those recognized incorrectly.}
	\label{result}
\end{figure}

We evaluate the effectiveness of our proposed components, including to utilize deformable convolutional layers, to use residual blocks and to resize the images to a larger size. Our model needs input images with size $200\times64$, and we also test the model with $100\times32$ which is applied in our baseline model. The comparisons are shown in Table \ref{ablation_components} and some of the recognition results are shown in Figure \ref{result}. The models are all trained with case insensitive mode and tested without lexicon. 

\begin{table}[htbp]
	\begin{center}
		\begin{tabular}{cccccccc}
			\hline\noalign{\smallskip}
			DConv     & ResBlock  &  Larger Size  & Total         & IC13         & IC15         & SVT           & IIIT5K        \\
			\noalign{\smallskip}\hline\noalign{\smallskip}
			&           &               & 64.8          & 86.2         & 65.3         & 78.4          & 85.2          \\
			\ding{52} &           &               & 68.6          & 88.7         & 70.8         & 78.3          & 90.4          \\
			&\ding{52}  &               & 68.2          & 88.9         & 69.5         & 78.7          & 91.4          \\
			&           &\ding{52}      & 65.2          & 87.4         & 65.6         & 78.1          & 87.4          \\
			\ding{52} &\ding{52}  &               & 68.7          & 89.0         & 69.9         & 78.9          & 90.9          \\
			\ding{52} &\ding{52}  &\ding{52}      & \textbf{70.3} &\textbf{89.7} &\textbf{72.2} & \textbf{79.4} & \textbf{92.2} \\
			\noalign{\smallskip}\hline
		\end{tabular}
	\end{center}
	\caption{Ablation Experiment Results. In the table, "DConv" means to utilize the deformable convolution layers, "ResBlock" means to add residual blocks into the network and "Larger Size" means to utilize $200\times 64$ as the size of input images.}
	\label{ablation_components}
\end{table}

From Table \ref{ablation_components}, we can observe that the adaptations of our model achieves a progress on the accuracy compared with the baseline model. And apparently, the deformable layers are chiefly effective on irregular images, but the effect is not significant when recognizing regular text. The residual blocks are mainly effective on regular images. And to utilize $200\times64$ as the input size dose not bring about significant improvements on our baseline networks. However, as some space for offsets is required, the larger input size is highly effective on our modified network. Finally, with all of the components, the model will reach the best performance. 

\subsubsection{The Impacts of Location of Deformable Layers}
\label{location_impact}
To demonstrate that it is best to utilize the deformbale convolutional layers as the fourth and fifth layers, ablation experiments with deformable layers at different positions have been involved. The results are shown in Table \ref{ablation_location}.
\begin{table}[htbp]
	\begin{center}
		\begin{tabular}{cccccc}
			\hline\noalign{\smallskip}
			Location    & Total        & IC13         & IC15        & SVT           & IIIT5K        \\
			\noalign{\smallskip}\hline\noalign{\smallskip}
			\{3\}         &  65.1        &   89.2       &  67.2       & 77.6          &  88.9         \\
			\{4\}         &  66.4        &   88.1       &  67.2       & 77.7          &  89.8         \\
			\{5\}         &  67.7        &   89.3       &  68.6       & 77.6          &  90.6         \\
			\{3,4\}       &  66.1        &   88.5       &  66.8       & 78.0          &  89.4         \\
			\{4,5\}       & \textbf{70.3}& \textbf{89.7}&\textbf{72.2}& \textbf{79.4 }& \textbf{92.2} \\
			\{3,5\}       &  64.4        &   88.3       &  64.4       & 74.6          &  89.1         \\
			\{3,4,5\}     &  63.8        &   87.3       &  64.6       & 75.0          &  88.1         \\ 
			\noalign{\smallskip}\hline
		\end{tabular}
	\end{center}
	\caption{Ablation Experiment Results. "Location" represents which convolutional layers are replaced with deformable ones. For instance, "\{3,5\}" means to replace the third and fifth layers.}
	\label{ablation_location}
\end{table}
In this table, it can be observed that when deeper layers are replaced with deformable ones, the network will achieve better performances. And to utilize two deformable layers is better than to use one, but when three layers are replaced, there is degeneration on accuracy. We consider that it is because too many deformable layers cause an over-fitting problem. According to the results, finally we choose to apply deformable layers in the fourth and fifth convolution.

\subsection{Comparative Evaluation}

\begin{table}[htbp]
	\begin{center}
		\begin{tabular}{cccccc}
			\hline\noalign{\smallskip}
			Methods                                       & Total       & IC13        & IC15        & SVT         & IIIT5K      \\
			\noalign{\smallskip}\hline\noalign{\smallskip}
			Shi \textsl{et al.}\cite{crnn}                &    -        & 86.7        &  -          & 80.8        & 78.2        \\
			Luo \textsl{et al.}\cite{moran}	              &    -        & 92.4        & 68.8        &\textbf{88.3}& 91.2        \\
			Liu \textsl{et al.}\cite{liu2016star}         &    -        & 89.1        &  -          & 83.6        & 83.3        \\
			Lyu \textsl{et al.}\cite{lyu2018mask}		  &  52.9       & 86.5        & 62.4        &  -          &  -          \\
			Sun \textsl{et al.}\cite{sun2018textnet}	  &  54.0       & 83.0        & 60.5        &  -          &  -          \\
			Shi \text{et al.}\cite{rare}                  &    -        & 88.6        &  -          & 81.9        & 81.9        \\
			Lee \textsl{et al.}\cite{r2am}                &    -        & 90.0        &  -          & 80.7        & 78.4        \\
			Cheng \textsl{et al.}\cite{aon}               &    -        &   -         & 68.2        & 82.8        & 87.0        \\
			Liu \textsl{et al.}\cite{liu2018char}         &    -        & 90.8        & 60.0        & 84.4        & 83.6        \\
			Liu \textsl{et al.}\cite{liu2018synthetically}&    -        &\textbf{94.0}&  -          & 87.1        & 89.4        \\
			Bissacco \textsl{et al.}\cite{bi2013photoocr} &    -        & 87.6        &  -          & 78.0        &  -          \\
			Wang \textsl{et al.}\cite{wang2017gated}      &    -        &   -         &  -          & 81.5        & 80.8        \\
			Jaderberg \textsl{et al.}\cite{jad2014deep}   &    -        & 81.8        &  -          & 71.7        &  -          \\
			Jaderberg \textsl{et al.}\cite{mjsynth2}      &    -        & 90.8        &  -          & 80.7        &  -          \\
			Tan \textsl{et al.}\cite{tan2014using}        &    -        &   -         &  -          & 80.1        & 81.7        \\
			Baseline                                      &  64.8       & 86.2        & 65.3        & 78.4        & 85.2        \\
			Ours                                          &\textbf{70.3}& 89.7        &\textbf{72.2}& 79.4        &\textbf{92.2}\\
			\noalign{\smallskip}\hline
		\end{tabular}
	\end{center}
	\caption{Experiment Results. In the table, "Baseline" represents the model which is trained by our training dataset using our baseline model without any modification. "-" represents that the authors did not test their model on the dataset and no result provided.}
	\label{comparisons}
\end{table} 

We make a few comparisons with other methods including \cite{crnn,moran,liu2016star,cheng2017focusing,lyu2018mask,sun2018textnet,rare,r2am,aon,liu2018char,liu2018synthetically,bi2013photoocr,wang2017gated,jad2014deep,mjsynth2,tan2014using}. All the results are reached without lexicon, which are shown in Table \ref{comparisons}. It is obvious that our method is effective while dealing with irregular text like those from TotalText \cite{totaltext} and ICDAR 2015 \cite{icdar15}. Though there are no significant improvements when treating regular text from ICDAR 2013 \cite{icdar13} and IIIT5K \cite{IIIT5K}, our proposed model still reaches a satisfactory performance. On SVT \cite{SVT}, our model do not achieve a high score, and we assert that is because many of the images are severely corrupted by noise and blur, or have very low resolutions. That is different from the images in the training dataset, which confuses the deformable layers. The deformable layers are not able to locate the text area learned from the training images, as our goal is to deal with irregular text but not blurred text.

\section{Conclusions}
\label{conclusions}

For dealing with irregular text images, modules for rectifying is usually easier to think of. But that is different from the way we read irregular text which should be to change and enhance our focus. In this work, we propose a method to recognize both regular and irregular text images through utilizing deformable convolutional layers to enable the ability of the network to change and enhance its focus. The model has reached a satisfactory performance, and no component for rectifying the images is applied. In the future, as more complicated recognition networks are available and attention mechanism can be involved, our goal is to design a system that is able to deal with images in which the text is in any orientation without preprocesses. 

\bibliographystyle{elsarticle-num} 
\bibliography{OSP_Latex_template}





\end{document}